\documentclass[twoside]{article}
\usepackage[accepted]{aistats2014}
\usepackage{verbatim}

\usepackage{graphicx} 
\usepackage[tight,sf,SF]{subfigure} 
\usepackage{multirow}

\usepackage{times}
\usepackage{graphicx} 
\usepackage{subfigure} 

\usepackage{wrapfig}

\usepackage{algorithm}
\usepackage{algorithmic}
\usepackage[colorinlistoftodos]{todonotes}

\usepackage{hyperref}
\hypersetup{
    unicode=false,          
    pdftoolbar=true,        
    pdfmenubar=true,        
    pdffitwindow=false,     
    pdfstartview={FitH},    
    colorlinks=true,       
   linkcolor=cyan,          
    citecolor=green,        
    filecolor=black,      
    urlcolor=black           
}

\usepackage{amssymb, amsmath}
\usepackage{dsfont}

\usepackage{color}

\newcommand\cut[1]{}

\newcommand{\R}{\mathds{R}}

\renewcommand{\vec}{\boldsymbol}
\newcommand{\mat}{\boldsymbol}
\newcommand{\E}{\mathds{E}}
\newcommand{\var}{\mathrm{var}}

\newcommand{\diag}{\mathrm{diag}}

\newcommand{\T}{^\top}
\newcommand{\inv}{^{-1}}

\newcommand{\gauss}[2]{\mathcal N(#1,#2)}
\newcommand{\gaussx}[3]{\mathcal{N}\big(#1\,|\,#2,#3\big)}

\newcommand{\idx}[1]{{(#1)}}

\newcommand{\LR}[2]{\mathrm{LR}_{#1}(#2)} 
\newcommand{\KL}[2]{\mathrm{KL}(#1||#2)} 

\definecolor{orange}{rgb}{1,0.5,0}
\definecolor{cyan}{rgb}{0.6, 0.2, 0.6}
\definecolor{magenta}{rgb}{1,0,1}
\definecolor{green2}{rgb}{0.2, 0.5, 0.2}

\usepackage{ifpdf}
\usepackage{array}
\usepackage{mdwmath}
\usepackage{mdwtab}

\usepackage{stfloats}

\begin{document}

%
\runningtitle{Large-Scale Gaussian Process Regression}

%

\twocolumn[

\aistatstitle{Hierarchical Mixture-of-Experts Model for \\  Large-Scale Gaussian Process Regression}

\aistatsauthor{ Jun Wei Ng \And Marc Peter Deisenroth }

\aistatsaddress{Department of Computing\\Imperial College London \And Department of Computing\\Imperial College London } ]

\begin{abstract}
We propose a practical and scalable Gaussian process model for large-scale nonlinear probabilistic regression. Our mixture-of-experts model is conceptually simple and hierarchically recombines computations for an overall approximation of a full Gaussian process. Closed-form and distributed computations allow for efficient and massive parallelisation while keeping the memory consumption small. Given sufficient computing resources, our model can handle arbitrarily large data sets,  without explicit sparse approximations. We provide strong experimental evidence that our model can be applied to large data sets of sizes far beyond millions. Hence, our model has the potential to lay the foundation for general large-scale Gaussian process research.
\end{abstract}

\section{Introduction}

Probabilistic Gaussian processes (GPs)~\cite{Rasmussen2006} are the method of choice for probabilistic regression: Their non-parametric nature allows for flexible modelling without specifying low-level assumptions (e.g., the degree of a polynomial) in advance. Moreover, for the standard Gaussian likelihood, inference can be performed in closed form in a principled way simply by applying Bayes' theorem. GPs have had substantial impact in various research areas, including geostatistics~\cite{Cressie1993}, optimisation~\cite{Jones1998,Brochu2009}, data visualisation~\cite{Lawrence2005}, robotics and reinforcement learning~\cite{Deisenroth2014}, spatio-temporal modelling~\cite{Luttinen2012}, and active learning~\cite{Krause2008}. A strength of the GP model is that it can be used  fairly reliably as a black-box function approximator, i.e., it produces reasonable predictions without manual parameter tuning.

A practical limitation of the GP model is its computational demand: In
standard implementations, training and predicting scale in
$\mathcal{O}(N^3)$ and $\mathcal{O}(N^2)$, respectively, where $N$ is
the size of the training data set. For large $N$ (e.g., $N>10,000$) we
often use sparse approximations~\cite{Williams2001,Quinonero-Candela2005,
  Hensman2013, Titsias2009, Lazaro-Gredilla2010, Shen2006}. Typically, these sparse approximations lower the computational burden by implicitly (or explicitly)  using a subset of the data. They scale GPs up to multiple tens or hundreds of thousands of data points. However, even with sparse approximations it is inconceivable to apply GPs to data set sizes of tens or hundreds of millions of data points. 

An alternative to sparse approximations is to distribute the computations by using local models. These local models typically require stationary kernels for a notion of ``distance'' and ``locality''. 
Shen et al.~\cite{Shen2006} used KD-trees to recursively partition the data space by means of a multi-resolution tree data structure, which allows to scale GPs up to multiple tens of thousands of training points. However, the approach proposed in~\cite{Shen2006} does not provide solutions for variance predictions and is limited to stationary kernels. Similarly, \cite{Nguyen-Tuong2009a} used a heuristic partitioning scheme based on the locality notion of stationary kernels for real-time mean predictions of GPs.
Along the lines of exploiting locality, mixture-of-experts (MoE) models~\cite{Jacobs1991} have been applied to GP regression~\cite{Meeds2006,Rasmussen2002,Yuan2009}. However, these models have not primarily been  used to speed up GP regression, but rather to allow for heteroscedasticity (input-dependent variations) and non-stationarity. Each local model possesses its own set of hyper-parameters to be optimised. Predictions are made by collecting the predictions of all local expert models, and weighing them using the responsibilities assigned by the gating network. In these MoE models, a Dirichlet process prior is placed on the multinomial ``responsibility'' vector of each local expert, which allows for a data-driven partitioning on the fly. Unfortunately, inference in these models is  computationally intractable, requiring MCMC sampling or variational inference to assign data points to each GP expert, a computationally demanding process.

Within the context of spatio-temporal models with $10^6$ data points, \cite{Luttinen2012} propose efficient inference that exploits computational advantages from both separable and compactly supported kernels, leading to very sparse kernel matrices. The authors propose approximate (efficient) sampling methods to deal with both noisy and missing data.

Recently, Gal et al.~\cite{Gal2014} proposed a distributed approach that scales variational sparse GPs further by exploiting distributed computations. In particular, they derive an exact re-parameterisation of the variational sparse GP model by Titsias~\cite{Titsias2009}, to update the variational parameters independently on different nodes. This is implemented within a Map-Reduce framework, and the corresponding architecture consists of a central node and many local nodes, i.e., a single-layer architecture.

In this paper, we follow an approach orthogonal to sparse approximations in order to \emph{scale full GPs to large data sets} by exploiting massive parallelisation. In particular, we propose a hierarchical mixture of experts model that distributes the computational load amongst a large set of independent computational units.  Our model recursively recombines computations by these independent units to an overall GP inference\slash training procedure. This idea is related to Tresp's Bayesian Committee Machine~\cite{Tresp2000}, which combines estimators. A key advantage of our model is that all computations can be performed analytically, i.e., no sampling is required. Given sufficient computing power our model can handle arbitrarily large data sets. In our experiments we demonstrate that our model can be applied easily to data sets of size $2^{24}\approx 1.7\times 10^7$, which exceeds the typical data set sizes sparse GPs deal with by orders of magnitude. However, even with limited computing resources, our model is practical in the sense that it can train a full GP with a million training points in less than half an hour on a laptop.

\section{Objective and Set-up}
We consider a regression setting $y=f(\vec x)+\epsilon$, where $\vec x\in\R^d$ and $\epsilon\sim\gauss{0}{\sigma_\epsilon^2}$ is i.i.d. Gaussian noise. The objective is to infer the latent function $f$ from a training data set $\mat X = \{\vec x_i\}_{i=1}^n, \vec y = [y_1, \dotsc, y_N]\T$. For small data set sizes $N$, a Gaussian process is one of the methods of choice for probabilistic non-parametric regression.

A Gaussian process is a non-parametric Bayesian model, which is often used for (small-scale) regression. A GP is defined as a collection of random variables, any finite number of which is Gaussian distributed. A GP is fully specified by a mean function and a covariance function (kernel). In this paper, we assume that the prior mean function is 0. Furthermore, we consider the Gaussian (squared exponential) kernel
\begin{align}
\hspace{-2.6mm}k(\vec x_i, \vec x_j) = \sigma_f^2\exp\big(-\tfrac{1}{2}(\vec x_i - \vec x_j)\T\mat\Lambda\inv(\vec x_i-\vec x_j)\big),
\end{align}
$\vec x_i,\vec x_j\in\R^D$, where $\sigma_f^2$ is the variance of the latent function $f$ and $\mat\Lambda = \diag(l_1^2,\dotsc,l_D^2)$ is the diagonal matrix of squared length-scales $l_i$, $i = 1,\dotsc,D$. Furthermore, we consider a Gaussian likelihood $p(y|f(\vec x))=\gauss{f(\vec x)}{\sigma_\epsilon^2}$ to account for the measurement noise. 

A GP is typically trained by finding hyper-parameters $\vec\theta = \{\sigma_f, l_i, \sigma_\epsilon\}$ that maximise the log-marginal likelihood~\cite{Rasmussen2006}, which is (up to a constant) given by
\begin{align}
\log p(\vec y|\mat X, \vec\theta) &\stackrel{.}{=} -\tfrac{1}{2}\big(\vec y(\mat K+\sigma_\epsilon^2\mat I)\inv\vec y + \log|\mat K+\sigma_\varepsilon^2\mat I|\big)\,.
\label{eq:log-marginal likelihood}
\end{align}
The computationally demanding computations in~\eqref{eq:log-marginal likelihood} are the inversion and the determinant of \mbox{$\mat K + \sigma_\epsilon^2\mat I$}, both of which scale in $\mathcal{O}(N^3)$ with a standard implementation.

For a given set of hyper-parameters $\vec\theta$, a training set $\mat X, \vec y$ and a test input $\vec x_*\in\R^D$, the GP posterior predictive distribution of the corresponding function value $f_* = f(\vec x_*)$ is Gaussian with mean and variance given by
\begin{align}
\E[f_*] &= m(\vec x_*) = \vec k_*\T(\mat K + \sigma_\varepsilon^2\mat I)\inv \vec y\,,\label{eq:mean GP}\\
\var[f_*] &= \sigma^2(\vec x_*) = k_{**} -\vec k_*\T(\mat K + \sigma_\varepsilon^2\mat I)\inv \vec k_*\,,\label{eq:var GP}
\end{align}
respectively, where $\vec k_* = k(\vec x_*, \mat X)$ and $k_{**}=k(\vec x_*, \vec x_*)$. When we cache $(\mat K + \sigma_\epsilon^2\mat I)\inv$ computing the mean and variance in~\eqref{eq:mean GP} and \eqref{eq:var GP} requires $\mathcal{O}(N)$ and $\mathcal{O}(N^2)$ computations, respectively.
%

For $N>10,000$ training and predicting become rather time-consuming procedures, which additionally require large amounts of memory, i.e., $\mathcal{O}(N(N+D))$. 

Our working hypothesis is that a standard GP can model the latent function $f$. However, due to the data set size $N$ the standard GP is not applicable.

Instead of a sparse approximation~\cite{Quinonero-Candela2005, Titsias2009}, we address both the computational and the memory issues of full GPs by distributing the computational and memory load to many individual computational units that only operate on subsets of the data. This results in an approximation of the full GP, but this approximation can be computed efficiently (time and memory) by exploiting massive parallelisation.

\section{Hierarchical Mixture-of-Experts Model}
\begin{figure*}[tb]
\subfigure[Single-layer model.]{
\includegraphics[height = 2.45cm]{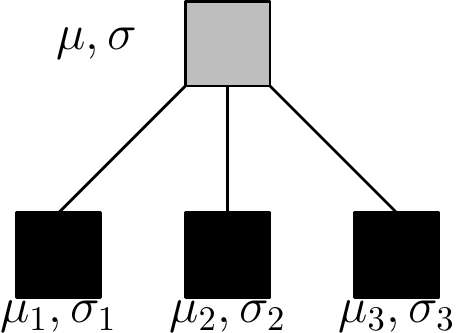}
\label{fig:1layer}
}
\hspace{15mm}
\subfigure[Two-layer model.]{
\includegraphics[height = 4cm]{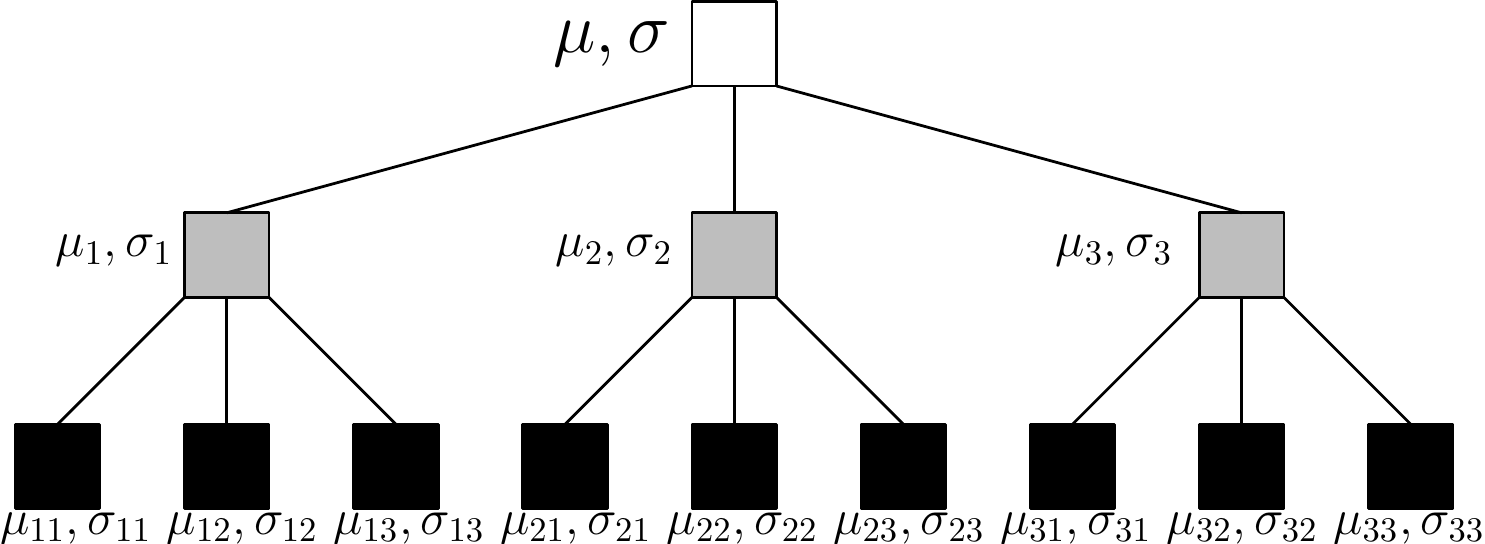}
\label{fig:2layer}
}
\caption{Hierarchical MoE model. Main computations are at the leaf nodes (black). All other nodes (linearly) recombine information from their direct children, allowing for an arbitrarily deep architecture.}
\end{figure*}

Consider a GP with a training data set $\mathcal{D} = \{\mat X, \vec y\}$.  We define a set $\mathcal{S}$ of $c$ subsets (not necessarily a partition) of the data set as $\mathcal{S} = \{\mathcal{D}^\idx{1},\dotsc,\mathcal{D}^\idx{c}\}$ where $\mathcal{D}^{(i)} = \{ \mat{X}^{(i)} , \vec{y}^{(i)} \}$. These subsets are from the full training set  $\mathcal{D}$, and we will use a GP on each of them as a (local) expert\footnote{The notion of ``locality'' is  misleading as our model does not require similarity measures induced by stationary kernels.}. Each of these local expert models computes means and variances conditioned on their respective training data\footnote{Both mean and variances are necessary for training and inference.}. These (local) predictions are recombined to a mean\slash variance by a parent node (see Fig.~\ref{fig:1layer}), which subsequently can play the role of an expert at the next level of the model architecture. Recursively applying these recombinations, our model results in a tree structured architecture with arbitrarily many layers, see Fig.~\ref{fig:2layer}.\footnote{We discuss different architecture choices in Section~\ref{sec:architectures}.}
In our model, all GPs at the leaves of this tree are trained jointly and share a single set of hyper-parameters $\vec\theta$.

\begin{figure*}[tb]
\centering
\includegraphics[width = 0.9\hsize]{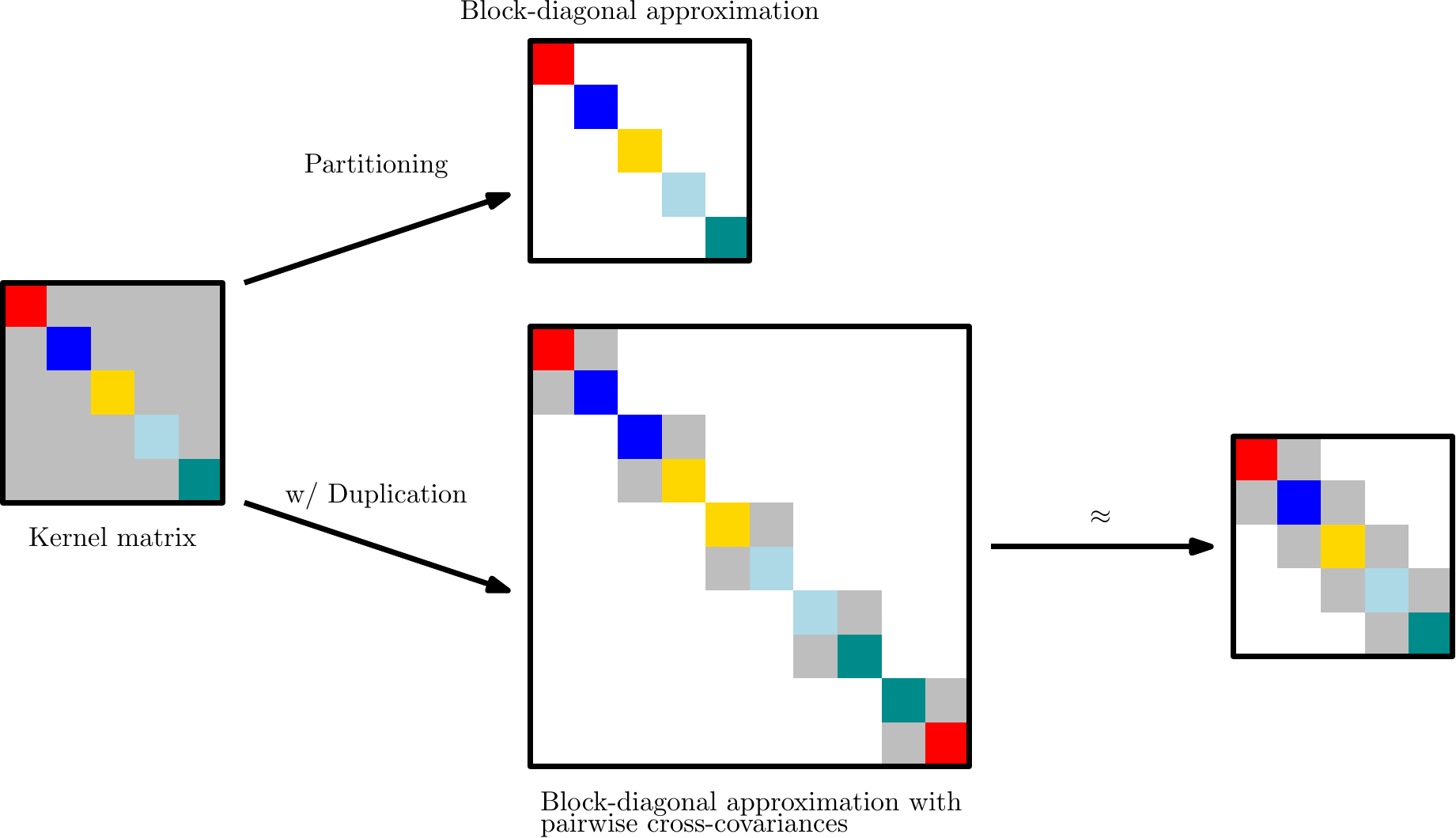}
\caption{In a single-layer MoE model, partitioning the data leads to a block-diagonal approximation of the kernel matrix (top path). By duplicating data points this clear separation between blocks is smoothed out (bottom path), and the effects of the independence assumption are mitigated.}
\label{fig:kernel matrix approx}
\end{figure*}

We train our model and make predictions under the assumption that each expert is independent of the other experts at the same level of the tree, which allows us to parallelise and distribute computations over independent processing units.
The assumption that the GPs at each level are independent effectively results in a GP with a block-diagonal covariance matrix, which can be efficiently inverted by distribution. If $\mathcal{S}$ is a partition of $\mathcal{D}$, this covariance matrix would be composed of the block-diagonal elements of the original GP's covariance matrix (with zeros on the off block-diagonals). However, with a partition, some information about the structure of the data is lost, which would otherwise be captured by the cross-covariance terms in the full GP.  Since our model aims to replicate a full GP we are interested in mitigating the effects of the independence assumption. We do so by sharing arts of the training set amongst multiple subsets in $\mathcal{S}$. Thereby, we account for the covariance effects between the points in $\mathcal{D}$ to some degree. This approach is illustrated in the bottom path of Fig.~\ref{fig:kernel matrix approx}, where parts of the training set are shared amongst individual nodes. Note that memory consumption can be kept constant since the training set is not modified (read-only access).

\subsection{Dividing the Data into Subsets}
Creating subsets of the training data and using each subset with a GP forms the basis of our hierarchical MoE GP (HGP) model, where we have divided the problem into a number of GPs, each using a smaller training data set.  This can be done recursively, further subdividing the problem until a desired training set size for the leaf-GPs is achieved.\footnote{
The data set sizes assigned to the leaves can be chosen depending on the computational resources available.
}
For an efficient implementation, the number $c$ of data subsets $\mathcal{D}^\idx{k}$ should correspond to a multiple of the number of computing units available. For disjoint data sets $\mathcal{D}^\idx{k}$, every leaf-GP would possess $N/c$ data points; if we allow for shared data sets this number scales linearly with the degree of sharing. For instance, if every data point appears in two local experts  in a single-layer model, we would have each $\mathcal{D}^\idx{k}$ of size $2N/c$.

There are various ways of assigning data points to the experts at the leaves of the tree in Fig.~\ref{fig:2layer}. For instance, random assignment is fast and can work quite well. Most of our results, however are based on a different approach: First, we use a KD-tree to recursively divide the input space into non-overlapping regions. We terminate the division when the required number of partitions is reached. Second, each region is then partitioned into $p$ disjoint groups of inputs. Third, we construct each data set $\mathcal{D}^\idx{k},\,k = 1,\dotsc, c$, for the local experts, such that it contains exactly one group from each region.  After this procedure, each data set $\mathcal{D}^\idx{k}$ will contain points across the entire input space, rather than being clustered in the same region in the input space. Note that neither method for assigning data points to GP experts relies on any locality notion induced by the kernel.

\subsection{Training}
We train the model by maximising a factorising approximation to the log-marginal likelihood in~\eqref{eq:log-marginal likelihood}, i.e.,
\begin{align}
\log p(\vec y| \mat X, \vec\theta) \approx \sum\nolimits_k \log p(\vec y^\idx{k}|\mat X^\idx{k}, \vec\theta)
\label{eq:approximate log-marginal likelihood}
\end{align}
with respect to the kernel hyper-parameter $\vec\theta$, which are shared amongst all individual GPs. The factorising approximation in~\eqref{eq:approximate log-marginal likelihood} is implied by our independence assumption of the individual (small) GP models. Each term in~\eqref{eq:approximate log-marginal likelihood} is given by 
\begin{align}
\hspace{-2mm}\log p(\vec y^\idx{k}|\mat X^\idx{k},\vec\theta) &= -\tfrac{1}{2}\vec y^\idx{k}(\mat K_{\vec\theta}^\idx{k} + \sigma_\epsilon^2\mat I)\inv\vec y^\idx{k} \nonumber\\
&\quad -\tfrac{1}{2}\log |\mat K_{\vec\theta}^\idx{k} + \sigma_\epsilon^2\mat I| + \text{const}
\label{eq:term in approx. LML}
\end{align}
and requires the inversion and determinant of $\mat K_{\vec\theta}^\idx{k} +\sigma_\epsilon^2\mat I$, where $\mat K_{\vec\theta}^\idx{k} = k(\mat X^\idx{k}, \mat X^\idx{k})$ is a $p\times p$ matrix, and $p$ is the size of the data set associated with the GP expert $k$. These computations can be performed in $\mathcal{O}(p^3)$ time with a standard implementation. Note that $p$ is significantly smaller than $N$, i.e., the  size of the full data set. The memory consumption is $\mathcal{O}(p^2 + pD)$ for each individual model.

Note that in~\eqref{eq:approximate log-marginal likelihood} the number of parameters to be optimised is relatively small since we do not consider additional variational parameters or inducing inputs that we optimise. The gradients of~\eqref{eq:approximate log-marginal likelihood} and~\eqref{eq:term in approx. LML} with respect to $\vec\theta$ can be computed in independently at all $k$ nodes, which allows for massive parallelisation and a significant speed-up of training compared to training a full GP.

\begin{figure}[tb]
\centering
 \includegraphics[width = \hsize]{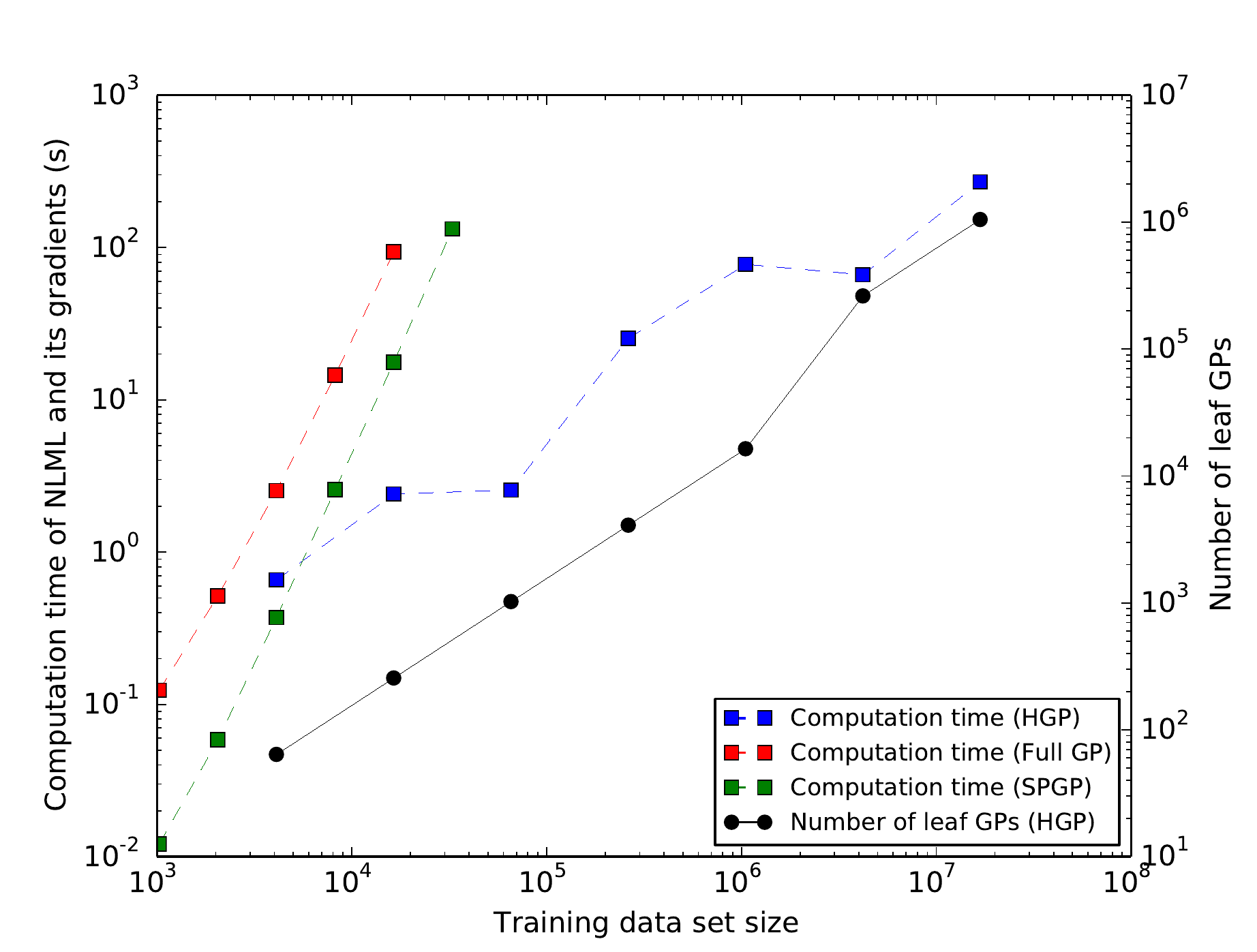}
\caption{Computing time for the log-marginal likelihood and its gradient with respect to the kernel hyper-parameters as a function of the size of the training data. The HGP scales favourably to large-scale data sets. With an increasing number of child-GPs (but fixed computational resources), the HGP scales to more than $10^7$ data points.}
\label{fig:hgp timings}
\end{figure}
%
%
To evaluate the training time for our GP model, we computed the amount of time required to compute the log-marginal likelihood and its gradients with respect to the kernel hyper-parameters. A typical optimisation procedure for the kernel hyper-parameters, e.g., conjugate gradients or (L)BFGS, requires these values. The full training time is proportional to the time it takes to compute the log-marginal likelihood and its gradient (it still depends on the number of line-searches). We chose a computer architecture of 64 nodes with 4 cores each. Furthermore, we chose a three-layer model with varying widths (branching factors). For data sets of $\leq 2^{20}$ data points the leaf nodes possessed 512 data points each, for data set sizes of $>2^{20}$, we chose the number of data points per node to be 128.

Fig.~\ref{fig:hgp timings} shows the time required for computing the log-marginal likelihood and its gradient with respect to the hyper-parameters. The horizontal axis shows the size of the training set (logarithmic scale), the left vertical axis shows the computation time in seconds (logarithmic scale) for our model (HGP, blue-dashed), a full GP (red-dashed) and a sparse GP with inducing inputs~\cite{Snelson2006} (green-dashed). For the sparse GP model, we chose the number $M$ of inducing inputs to be 10\% of the size of the training set, i.e., the computation time is of the order of $\mathcal{O}(NM^2)=\mathcal{O}(N^3/100)$, which offsets the curve of the full GP. Taking even fewer inducing inputs (e.g., 1\% or 0.1\% of the data) would push the sparse approximation towards multiple-hundred thousand data points. However, this can only be done if the data set possesses a high degree of redundancy. The right vertical axis shows the number of leaf GPs (black-solid), i.e., the number of GP experts amongst which we distribute the computation.  While the training time of the full GP reaches impractical number at data set sizes of about 10,000, the sparse GP model can be reasonably trained up to 50,000 data points.\footnote{In this comparison we did not include any computational overhead for selecting the inducing inputs (either as variational parameters or as free parameters to be optimised), which is often non-negligible.} The computational time required for the HGP to compute the marginal likelihood and gradients is significantly lower than that of the full GP, and we scaled it up to $2^{24} \approx 1.7 \times 10^7$ training data points, which required about the same amount of time ($\approx 230$\,s) for training a full GP with $2^{14}\times 10^4$ and a sparse GP with $2^{15}\approx 3.2\times 10^4$ data points. The figure shows that for any problem size, we are able to find an architecture that allows us to compute the marginal likelihood (hence, train the model) within a feasible amount of time.

Even if a big computing infrastructure is not available, our model is useful in practice: We performed a full GP training cycle (which includes many evaluations of the log-marginal likelihood and its gradient) with $10^6$ data points on a standard laptop in about 20 minutes. This is also a clear indicator that the memory consumption of the HGP is relatively small.




\subsection{Predictions/Inference}
The predictive distribution is computed by an iterative recombination of the computations at the leaf-GPs. In particular, the parent nodes compute
\begin{align}
p(y_*|\vec x_*, \mat X, \vec y) &\propto \prod\nolimits_k p(y_*|\vec x_*, \mat X^\idx{k}, \vec y^\idx{k}) \label{eq:predictive_likelihood}\\
&\propto\gaussx{y_*}{\mu_*}{\sigma_*^2}\,,\label{eq:predictive_distribution} 
\end{align}
as the product of all Gaussians passed on by the children. The resulting distribution is also Gaussian with
mean $\mu_*$ and variance $\sigma_*^2$, where
\begin{align*}
\mu_* = \sigma_*^{2}\sum\nolimits_k \frac{\mu_k(\vec x_*)}{\sigma_k^2(\vec x_*)}\,,\quad
\sigma_*^2 = \left(\sum\nolimits_k \sigma_k^{-2}(\vec x_*)\right)^{-1}.
\end{align*}
Both $\mu_*$ and $ \sigma_*^2$ are computed analytically. We exploit the distributed architecture of our model to ensure small computational overhead. The mean of the HGP predictive distribution can be written as a weighted sum of the means of the child-GPs' predictive distributions. $\mu_* = \sum\nolimits_k w_k \mu_k(\vec x_*)$ where
$w_k = \sigma_k^{-2}(\vec x_*)/\sum\nolimits_j \sigma_j^{-2}(\vec x_*)$. The weights on the child-GPs' predictions are proportional to the inverse variances of their prediction, which allows more accurate predictions (lower variances) to have bigger weights in the final prediction, and less accurate predictions (higher variances) to have weights closer to zero. This, in general, allows the HGP to remain effective across various methods of assigning data points to the child-GPs.

\subsection{Architecture Choice}
\label{sec:proofs}
Thus far, we have described the single-level version of the HGP model, where
the child-GPs are standard GPs. Since the HGP possesses the same ``interface''
(marginal likelihood, predictive distribution) as a standard GP, we can use
HGPs themselves as the child-GPs of an HGP. This can be done recursively to build up a tree
structure of arbitrary depth and width. 

In the following, we show that a multi-level HGP is
equivalent to a single-level HGP if its children (experts) are the leaf-GPs (experts) of a
multi-level HGP. For this, we show that training and prediction are identical in both models.

\paragraph{Training} In \eqref{eq:approximate log-marginal likelihood}, we
expressed the log-marginal likelihood of the HGP as a sum of the log marginal
likelihoods of its child-GPs. If the child-GPs themselves are also HGPs (``child-HGPs''),
then this sum can be expanded and expressed as the sum of the log-marginal
likelihood of the child-GPs of the child-HGPs. This generalises to HGPs of
an arbitrary number of levels, and we can always write the log-marginal likelihood
of a HGP as a sum of the log-marginal likelihood of its leaf GPs (experts):
\begin{align}
\log p(\vec y| \mat X, \vec\theta) &\approx \sum\nolimits_k \log p(\vec y^\idx{k}|\mat X^\idx{k}, \vec\theta) \nonumber \\
&\approx \sum\nolimits_k \sum\nolimits_{i_k} \log p(\vec y^\idx{i_k}|\mat X^\idx{i_k}, \vec\theta) \nonumber \\
&\approx \cdots \nonumber \nonumber \\
&\approx \sum\nolimits_{l \in leaves} \log p(\vec y^\idx{l}|\mat X^\idx{l}, \vec\theta)
\label{eq:hgp_lik_expansion}
\end{align}
Equation~\eqref{eq:hgp_lik_expansion} shows that the log-marginal likelihood of a multi-level HGP
is the sum of the log marginal-likelihoods of its leaves, which is equivalent
to the log-marginal likelihood of a single-level HGP with the (multi-level HGP's) leaves as its child-GPs (experts). Hence, the structure of the HGP above the leaves has no effect on the computation of the log-marginal likelihood.

\paragraph{Prediction}
We now show that the predictions of a multi-level HGP and a single-level HGP are identical. The product in \eqref{eq:predictive_likelihood} can be factorised\footnote{This is not strictly a factorisation, since it is only proportional to \eqref{eq:predictive_likelihood}. While it is not crucial to our application, we can easily recover the normalising constant by integration.} into
$\prod\nolimits_{l \in leaves} p(y_*|\vec x_*, \mat X^\idx{l}, \vec y^\idx{l})$,
a product of terms involving only the leaf-GPs (experts) and not containing terms relating to the intermediate levels.

It is, however, not immediately obvious that the predictive distribution in
\eqref{eq:predictive_distribution} is equivalent to the predictive distribution of a single-level HGP with the same leaves. To show this, we provide a simple proof that the resulting distribution of the product of an arbitrary number of Gaussians, which are in turn the product of Gaussians (we refer to them as the ``sub-Gaussians''), is Gaussian and equivalent to the distribution resulting from the product of all the sub-Gaussians.

The product of Gaussians is proportional to a Gaussian, i.e.,  
$\prod\nolimits_k \gaussx{x}{\mu_k}{\sigma_k^2} \propto
\gaussx{x}{\mu_*}{\sigma_*^2}
$ where
$\mu_* = \sigma_*^2\sum\nolimits_k \mu_k\sigma_k^{-2}$
and $\sigma_*^2 = (\sum\nolimits_k \sigma_k^{-2})^{-1} $.
Suppose each of the component Gaussians are themselves product of Gaussians.
That is,
$\gaussx{x}{\mu_k}{\sigma_k^2}
\propto
\prod\nolimits_{i_k} \gaussx{x}{\mu_{i_k}}{\sigma_{i_k}^2}
$ and
$\mu_k = \sigma_k^2\sum\nolimits_{i_k} \mu_{i_k}\sigma_{i_k}^{-2}$
and $\sigma_k^2 = (\sum\nolimits_{i_k} \sigma_{i_k}^{-2})^{-1} $. Then
\begin{align}
\mu_* &= \sigma_*^2\sum\nolimits_k \mu_k\sigma_k^{-2}
= \sigma_*^2\sum\nolimits_k \sum\nolimits_{i_k} \mu_{i_k}\sigma_{i_k}^{-2}\,,\\
\sigma_*^2 &= 
\big(\sum\nolimits_k \sigma_k^{-2}\big)^{-1} =
\big(\sum\nolimits_k \sum\nolimits_{i_k} \sigma_{i_k}^{-2} \big)^{-1}\,,
\end{align}
(where $i_k$ are the indices corresponding to the children of the child GPs, and so on)
i.e., the distribution from the product of the Gaussian distributions is equivalent to the distribution from the product of the sub-Gaussians.
This result generalises to any number of levels of Gaussian products (if the sub-Gaussians are derived as products of ``sub-sub-Gaussians'' we apply the above again), and completes our proof for the equivalence of a multi-level HGP and a single-level HGP with the same leaves (experts).

Therefore, mathematically it does not matter whether to choose a shallow or deep architecture if the leaf GPs (experts) are the same. However, a multi-level HGP still makes sense from a computational point of view.

\begin{figure}
\includegraphics[width = \hsize]{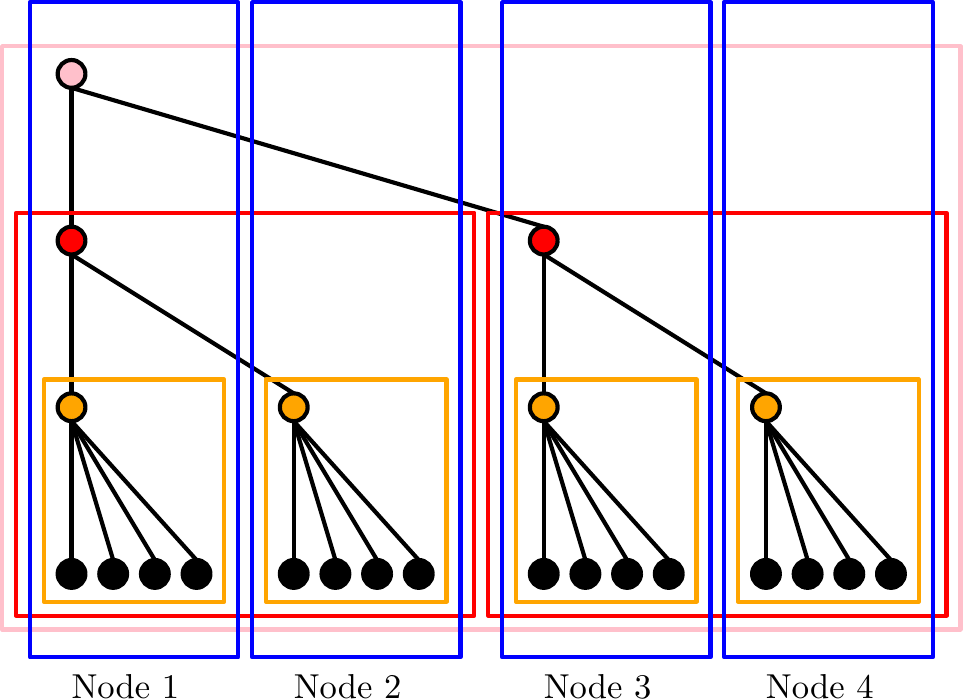}
\caption{The flexibility of choosing amongst equivalent architectures enables the HGP to evenly distribute computational (and network communication) work. Each blue vertical rectangle represents one distributed computing unit while each coloured node denotes an HGP. The coloured rectangles represent the overall responsibility of the corresponding coloured nodes. The overlap between the coloured rectangles and the blue represent the computing resources available for the computational work related to a particular HGP. The main computations are performed at the leaf-experts (black).}
\label{fig:arch}
\end{figure}

\paragraph{Multi-level HGPs}
Given the same leaf-GPs, the depth of the HGP has no effect on the model, for both training and prediction, as shown in Section~\ref{sec:proofs}. 
%
Although it is mathematically not necessary to construct an HGP of more than one level, in practice, a multi-level HGP allows us to fully utilise a given set of distributed computing hardware. To implement a single-level HGP on a distributed system, we have one ``master'' node, which is responsible for the computational work of the HGP (combining the outputs of the child-GPs). The computational work of the child-GPs is distributed evenly across the other ``slave'' nodes. Such a set-up imposes a heavy communication and computational load on the master node, since it has to manage its communication with all slave nodes, and perform the computations required for combining the child-GPs on its own (during which the slave nodes will idle). This is not an optimal use of resources, and we exploit the fact that the HGP model is invariant to the presence of intermediate layers to propose a better solution, which is illustrated in Fig.~\ref{fig:arch}. Starting from the top of the HGP tree, divide the number of computational nodes available into $c$ groups where $c$ is the number of child-GPs that the HGP  possesses, and assign each child-GP/child-HGP to one group. We do this recursively until we reach the leaves of the HGP or until there is only a single node available to the HGP. This approach leads to a more uniform distribution of network communication and computational load amongst all nodes.

\begin{figure*}[tb]
\begin{center}
\includegraphics[width=0.76\hsize]{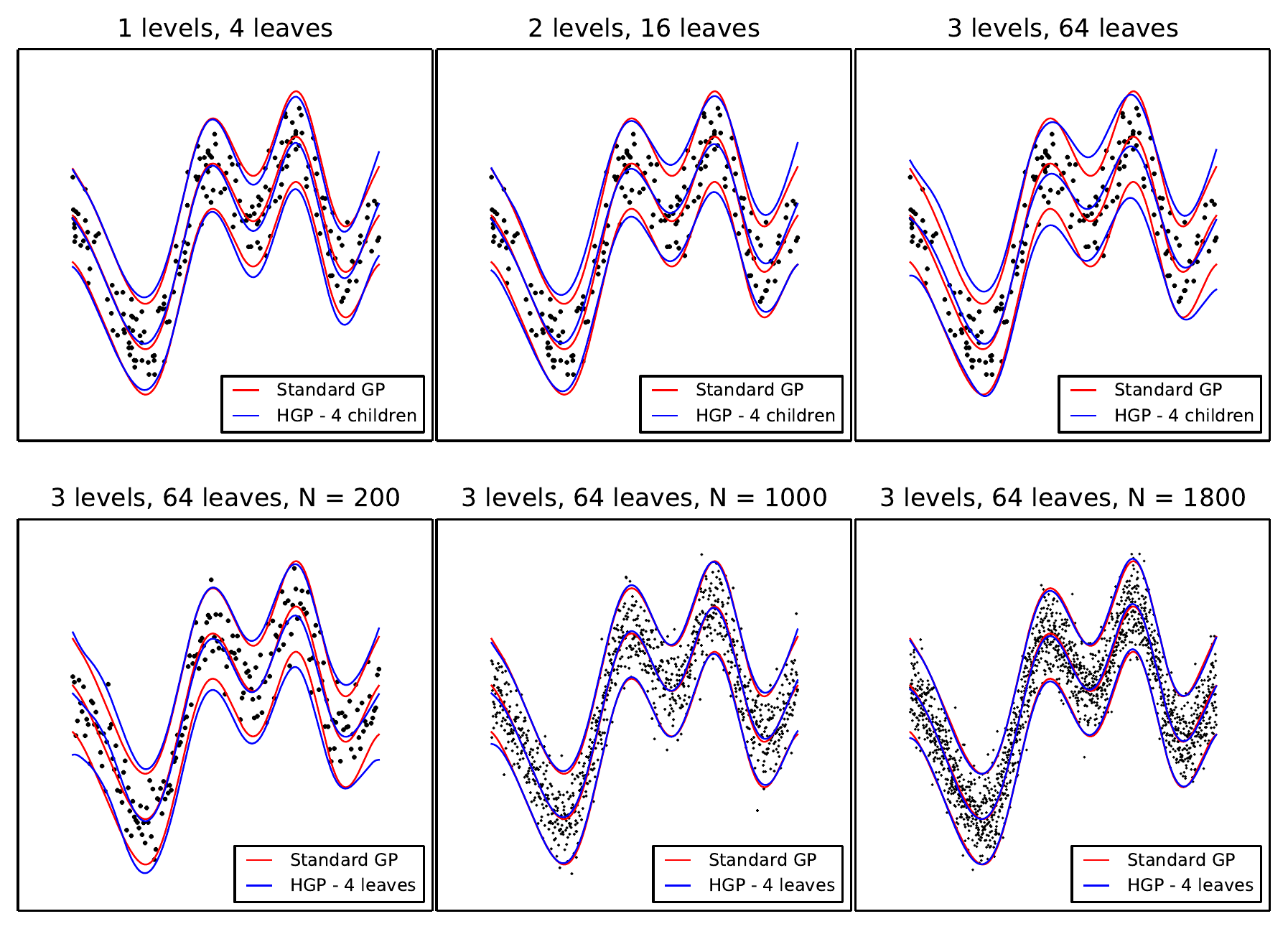}
\caption{Top row: Comparison of HGPs (blue lines) with varying depths (hence, number of experts) with a ground truth GP (red lines). The mean functions and corresponding $2\sigma$ predictive intervals are shown. The model accuracy decreases with the number of experts. Bottom row: For a fixed depth (and number of experts), the HGP model becomes more and more similar to the ground truth GP model with an increasing number of data points.}
\label{fig:depth_ndata}
\end{center}
\end{figure*}

\paragraph{Number of Experts}
Fig.~\ref{fig:depth_ndata} (top row) illustrates the effect of the number of leaf-GPs (experts) on the accuracy of the HGP. We constructed 3 HGPs with 1, 2, and 3 levels (4, 16, and 64 experts), respectively, on a training data set of size 200. This resulted in each HGP having experts with data sets  of sizes 100, 50, and 25, respectively. As the number of experts increases, the accuracy decreases. Especially with 64 leaves, no expert has enough data points to model the underlying latent function reasonably well. However, with more training data the HGP with 64 experts recovers the prediction accuracy of the full GP.

%

\subsection{Implementation: True Concurrency in Python}
A known issue of the CPython interpreter, which we use in our implementation, is the lack of true concurrency using the in-built threading library. Due to
the \emph{Global Interpreter Lock} (GIL, which is implemented in the interpreter because Python's memory management is not thread safe), only a single thread of Python code can be executed at any point in time. Therefore, the use of threads in the Python context only provides logical concurrency in terms of the flow of programs, but not true simultaneous computations. 

There exists a workaround for the true concurrency problem in Python, via the use of processes instead of threads to perform simultaneous computations. In the POSIX model, threads are lightweight units of computations belonging to the same process, thus, sharing the same memory space. Processes have their own memory space and come with increased system overheads compared to threads. However, on Linux (which we use for this implementation), the creation of duplicate processes (forking) does not incur large memory costs since Linux implements a copy-on-write model. This means that when a process forks into two, the memory is not copied, unless the new process attempts to modify it. In the context of our implementation, we make no modification to the training data, which is shared amongst all child-GPs. In terms of the memory  usage, each child-GP only needs to compute its own kernel matrix and the corresponding Jacobian matrix per hyper-parameter, which have no interaction with any other child-GP. Therefore, computing each child-GP using a separate process does not incur any large, redundant memory costs that would not be present in a true concurrency model implemented by native threads.



\section{Experimental Results}

In this section, we apply our model to two real-world data sets. For both the full GP (if applicable) and the HGP model we optimised the kernel hyper-parameters by maximising the log-marginal likelihood using BFGS. 
In all following experiments, we used a single standard desktop computer with four cores (16 GB RAM, Intel Core-i7). 

\subsection{Robotic-Arm Data}
We applied our HGP model to the kin40k data set~\cite{kin40k}. The kin40k data set is generated with a simulation of the forward dynamics of an 8-link all-revolute robot arm. The task in the data set is to predict the distance of the end-effector from a target using joint positions and twist angles. The training set size is 10,000 (i.e., we can still train a full GP on it for comparison), the test set size is 30,000, and the input dimension is 8. We trained our model with various architectures (fixed branching factor), ranging from a single-layer model with four experts to a model with 7 layers and $4^7=16,384$ experts. We chose different architectures to assess the trade-off between computation time and accuracy with respect to the full GP.

\begin{table*}
\caption{Overview of the kin40K-data set results.}
\label{tab:kin40K}
\begin{center}
\begin{tabular}{c|c|c|c|c|c} 
\bf \shortstack{\-\\Model\\\-}  &
\bf \shortstack{\-\\Number of Levels\\(HGP)} &
\bf \shortstack{\-\\Number of Leaves\\(HGP)} &
\bf \shortstack{\-\\Training Time (s)\\ per BFGS Iteration}  &
\bf \shortstack{\-\\Data Points\\per Leaf} &
\bf \shortstack{\-\\Likelihood\\ Ratio}
\\ \hline
GP (ground truth) & - & - & 218.5 &  10,000  & 1 \\ \hline
\multirow{7}{*}{HGP}
       & 1 & $4  $  & 75.6 &  5,000   & 0.992 \\ \cline{2-6}
       & 2 & $4^2=16$  & 56.5 &  2,500   & 0.978 \\ \cline{2-6}
       & 3 & $4^3=64$  & 52.0 &  1,250   & 0.956 \\ \cline{2-6}
       & 4 & $4^4=256$  & 49.4 &  625   & 0.909 \\ \cline{2-6}
       & 5 & $4^5=1024$  & 32.2 &  313 & 0.875 \\ \cline{2-6}
       & 6 & $4^6=4096$  & 17.1  & 157  & 0.834 \\ \cline{2-6}
       & 7 & $4^7=16384$  & 22.0 &  79  & 0.815 
\end{tabular}
\end{center}
\end{table*}
Table~\ref{tab:kin40K} summarises the results. We report the training time per BFGS iteration (all models required 50--70 iterations), the number of data points per computational unit, and the likelihood ratio $\LR{GP}{HGP}$, which tells us how close our model is to the full GP. The likelihood ratio $\LR{G_1}{G_2}$ of two distributions $G_1$ and $G_2$ is defined as
\begin{align*}
\LR{G_1}{G_2} &:=
\prod_{i=1}^N \frac{p(y_i|G_2)}{p(y_i|G_1)} \stackrel{N\to\infty}{\longrightarrow} \exp{\big( - \KL{G_1}{G_2}  \big)}
\end{align*}
where $y_i\sim G_1$ (see supplementary material for the proof).
The basic single-level HGP with only four experts was able to achieve very similar results in a significantly shorter amount of time. The performance of the HGP decreased with increasing depth since the number of data points per expert becomes too small (note that the input space is 8D) as discussed in Fig.~\ref{fig:depth_ndata}.

\subsection{Airline Delays (US Flight Data)}
We considered a data set reporting flight arrival and departure times for every commercial fight in the US from January to April 2008. This dataset contains extensive information about almost 2 million  flights, including the delay (in minutes) in  reaching the destination. For this data set, we followed the procedure described in~\cite{Hensman2013}\footnote{The data set can be obtained from \url{http://stat-computing.org/dataexpo/2009/}. Thanks to J Hensman for the availability of the pre-processing script.}: We randomly selected 800,000 data points from which we used a random subset of 700,000 samples to train the model and 100,000 to test it.   We chose the same eight input variables $\vec x$ as in~\cite{Hensman2013}: age of the aircraft, distance that needs to be covered, airtime, departure and arrival times, day of the week and month, month. This data set has been evaluated in~\cite{Hensman2013, Gal2014}, both of which use sparse variational GP methods to deal with this training set size. We applied our HGP model, using data duplication (each training instance is used by two experts) and 1,400 experts  with 1,000 data points each. Data was assigned randomly to the expert GPs. 

We repeated the experiment four times. The average training time was 35.6 minutes and 14 BFGS iterations.
Table~\ref{tab:airline} reports the average RMSE values for predicting airline delays.
\begin{table}[tb]
\centering
\caption{Average RMSE values for predicting airline delays (700K training data, 100K test data).}
\label{tab:airline}
\begin{tabular}{c|c|c}
SVGP~\cite{Hensman2013} & Dist SVGP~\cite{Gal2014} & HGP\\
\hline
 33.00 & 32.95 & \textbf{27.45}
\end{tabular}
\end{table}
Table~\ref{tab:airline} also relates our results for predicting airline delays to the corresponding results reported in~\cite{Gal2014}, where 100 inducing points were used for the sparse variational GP (SVGP)~\cite{Hensman2013} and for the distributed sparse variational GP (Dist SVGP)~\cite{Gal2014}, which are in line  with the results reported in~\cite{Hensman2013}. Compared to the sparse variational GP methods, our HGP achieves a substantially better predictive performance. Additionally, the HGP converged after a few tens of iterations, whereas the sparse variational GP methods~\cite{Hensman2013,Gal2014} require hundreds or thousands of iterations.

\section{Discussion}

Our approach to scaling up GPs is conceptually straightforward and practical: It recursively recombines computations by independent lower-level experts to an overall prediction. Unlike any other approach to scaling GPs to large data sets our model is not an explicit sparse approximation of the full GP.
Therefore, the leaf nodes still perform full GP computations, i.e., their computations scale cubically in the number of data points. However, the number of data points at each leaf can be controlled by adjusting the number of leaves.

In the limit of a single expert, our hierarchical GP model is equivalent to a standard GP. Additionally, even with more than a single expert, our hierarchical mixture-of-experts model is still a Gaussian process: Any finite number of function values is Gaussian distributed, although we make an implicit (smoothed out) block-diagonal approximation of the covariance matrix. Note that the Deep GP model~\cite{Damianou2013} is actually not a GP. 

In our model, the kernel hyper-parameters are shared amongst all local GP experts. This makes sense as our objective is to reproduce a full ``vanilla'' GP, which, for practical reasons (size of training set) cannot be applied to the problem. Shared hyper-parameters also do not suffer much from overfitting problems: Even if a local expert fits a poor model, its (wrong/biased) gradient only has a small contribution if the number of local models is high. 
Training our model is relatively easy: Besides the shared GP hyper-parameters there are no additional parameters, such as inducing inputs~\cite{Gal2014, Titsias2009, Snelson2006}, i.e.,  compared to these approaches it is less likely to end up in local optima.

The main purpose of our model is to scale up the vanilla GP by distributing computational load. Therefore, all kinds of variations that are conceivable for standard GPs could also be implemented in our model. In particular, this includes classification, sparse approximations, heteroscedastic models, and non-Gaussian likelihoods. Note that these models would only be necessary at the level of the leaf GPs: All other computations are (linear) recombinations of the computations at the leaves.

Compared to other mixture-of-experts models, we chose the simplifying assumption that we know the number of experts, which in our case corresponds to the individual computing units. Thus, we do not need a Dirichlet process prior over partitions and, hence, sampling methods, which dramatically simplifies (and speeds up) training and inference in our model.

\section{Conclusion and Future Work}
We presented a conceptually straightforward, but effective, hierarchical  model that allows to scale probabilistic Gaussian processes to (in principle) arbitrarily large data sets. The key idea behind our model is to massively parallelise computations by distributing them amongst independent computational units. A recursive (and closed-form) recombination of these independent computations results in a practical hierarchical mixture-of-GP-experts model that is both computationally and memory efficient. Compared to the most recent sparse GP approximations, our model performs very well, learns fast, requires little memory, and does not suffer from high-dimensional variational optimisation of inducing inputs. We have demonstrated that our model scales well to large data sets: (a) Training a GP with a million data points takes less than 30 minutes on a laptop, (b) with more computing power training a GP with more than $10^7$ data points can be done in a few hours.

The model presented in this paper lays the foundation for a variety of future work in the context of Gaussian processes, including classification, non-Gaussian likelihoods in regression, and the combination with sparse GP methods (for really large data sets and limited computing power) at the level of the leaf nodes of our mixture-of-experts model.

\subsubsection*{Acknowledgements}
MPD was supported by an Imperial College Junior Research Fellowship.

\bibliographystyle{abbrv}
\bibliography{hgp_literature.bib}

\begin{thebibliography}{10}

\bibitem{kin40k}
Kin {Family of Datasets}.

\bibitem{Brochu2009}
E.~Brochu, V.~M. Cora, and N.~{de Freitas}.
\newblock A {Tutorial on Bayesian Optimization of Expensive Cost Functions,
  with Application to Active User Modeling and Hierarchical Reinforcement
  Learning}.
\newblock Technical Report TR-2009-023, Department of Computer Science,
  University of British Columbia, 2009.

\bibitem{Cressie1993}
N.~A.~C. Cressie.
\newblock {\em Statistics for Spatial Data}.
\newblock Wiley-Interscience, 1993.

\bibitem{Damianou2013}
A.~Damianou and N.~D. Lawrence.
\newblock Deep {Gaussian Processes}.
\newblock In {\em Proceedings of the International Conference on Artificial
  Intelligence and Statistics}, volume~31 of {\em JMLR W\&CP}, pages 207--215,
  2013.

\bibitem{Deisenroth2014}
M.~P. Deisenroth, D.~Fox, and C.~E. Rasmussen.
\newblock Gaussian {Processes for Data-Efficient Learning in Robotics and
  Control}.
\newblock {\em IEEE Transactions on Pattern Analysis and Machine Intelligence},
  36, 2014.

\bibitem{Gal2014}
Y.~Gal, M.~van~der Wilk, and C.~E. Rasmussen.
\newblock Distributed {Variational Inference in Sparse Gaussian Process
  Regression and Latent Variable Models}.
\newblock In {\em Advances in Neural Information Processing Systems}. 2014.

\bibitem{Hensman2013}
J.~Hensman, N.~Fusi, and N.~D. Lawrence.
\newblock Gaussian {Processes for Big Data}.
\newblock In {\em Proceedings of the Conference on Uncertainty in Artificial
  Intelligence}. AUAI Press, 2013.

\bibitem{Jacobs1991}
R.~A. Jacobs, M.~I. Jordan, S.~J. Nowlan, and G.~E. Hinton.
\newblock Adaptive {Mixtures of Local Experts}.
\newblock {\em Neural Computation}, 3:79--87, 1991.

\bibitem{Jones1998}
D.~R. Jones, M.~Schonlau, and W.~J. Welch.
\newblock Efficient {Global Optimization of Expensive Black-Box Functions}.
\newblock {\em Journal of Global Optimization}, 13(4):455--492, December 1998.

\bibitem{Krause2008}
A.~Krause, A.~Singh, and C.~Guestrin.
\newblock Near-{Optimal Sensor Placements in Gaussian Processes: Theory,
  Efficient Algorithms and Empirical Studies}.
\newblock {\em Journal of Machine Learning Research}, 9:235--284, February
  2008.

\bibitem{Lawrence2005}
N.~Lawrence.
\newblock Probabilistic {N}on-linear {P}rincipal {C}omponent {A}nalysis with
  {G}aussian {P}rocess {L}atent {V}ariable {M}odels.
\newblock {\em Journal of Machine Learning Research}, 6:1783--1816, November
  2005.

\bibitem{Lazaro-Gredilla2010}
M.~L\'{a}zaro-Gredilla, J.~Quiñonero-Candela, C.~E. Rasmussen, and A.~R.
  Figueiras-Vidal.
\newblock Sparse {Spectrum Gaussian Process Regression}.
\newblock {\em Journal of Machine Learning Research}, 11:1865--1881, June 2010.

\bibitem{Luttinen2012}
J.~Luttinen and A.~Ilin.
\newblock Efficient {Gaussian Process Inference for Short-Scale Spatio-Temporal
  Modeling}.
\newblock In {\em Proceedings of the International Conference on Artificial
  Intelligence and Statistics}, volume~22 of {\em JMLR W\&CP}, pages 741--750,
  2012.

\bibitem{Meeds2006}
E.~Meeds and S.~Osindero.
\newblock An {Alternative Infinite Mixture of Gaussian Process Experts}.
\newblock In {\em Advances in Neural Information Processing Systems}. The MIT
  Press, 2006.

\bibitem{Nguyen-Tuong2009a}
D.~Nguyen-Tuong, M.~Seeger, and J.~Peters.
\newblock Model {Learning with Local Gaussian Process Regression}.
\newblock {\em Advanced Robotics}, 23(15):2015--2034, 2009.

\bibitem{Quinonero-Candela2005}
J.~Qui{\~n}onero-Candela and C.~E. Rasmussen.
\newblock A {Unifying View of Sparse Approximate Gaussian Process Regression}.
\newblock {\em Journal of Machine Learning Research}, 6(2):1939--1960, 2005.

\bibitem{Rasmussen2002}
C.~E. Rasmussen and Z.~Ghahramani.
\newblock Infinite {Mixtures of Gaussian Process Experts}.
\newblock In {\em Advances in Neural Information Processing Systems}, pages
  881--888. The MIT Press, 2002.

\bibitem{Rasmussen2006}
C.~E. Rasmussen and C.~K.~I. Williams.
\newblock {\em Gaussian {Processes for Machine Learning}}.
\newblock The MIT Press, Cambridge, MA, USA, 2006.

\bibitem{Shen2006}
Y.~Shen, A.~Y. Ng, and M.~Seeger.
\newblock Fast {Gaussian Process Regression Using KD-Trees}.
\newblock In {\em Advances in Neural Information Processing Systems}, 2006.

\bibitem{Snelson2006}
E.~Snelson and Z.~Ghahramani.
\newblock Sparse {Gaussian Processes using Pseudo-inputs}.
\newblock In {\em Advances in {Neural Information Processing Systems} 18},
  pages 1257--1264. The MIT Press, Cambridge, MA, USA, 2006.

\bibitem{Titsias2009}
M.~K. Titsias.
\newblock Variational {Learning of Inducing Variables in Sparse Gaussian
  Processes}.
\newblock In {\em Proceedings of the Twelfth International Conference on
  Artificial Intelligence and Statistics}, 2009.

\bibitem{Tresp2000}
V.~Tresp.
\newblock A {Bayesian Committee Machine}.
\newblock {\em Neural Computation}, 12(11):2719--2741, 2000.

\bibitem{Williams2001}
C.~K. Williams and M.~Seeger.
\newblock Using the {Nystr{\"o}m Method to Speed up Kernel Machines}.
\newblock In {\em Advances in Neural Information Processing Systems}, pages
  682--688, 2001.

\bibitem{Yuan2009}
C.~Yuan and C.~Neubauer.
\newblock Variational {Mixture of Gaussian Process Experts}.
\newblock In {\em Advances in Neural Information Processing Systems}, pages
  1897--1904, 2009.

\end{thebibliography}

\appendix

\end{document}


%

%

\twocolumn[

\aistatstitle{Appendix}

\aistatsauthor{ Anonymous Author 1 \And Anonymous Author 2 \And Anonymous Author 3 }

\aistatsaddress{ Unknown Institution 1 \And Unknown Institution 2 \And Unknown Institution 3 } ]

\section{Likelihood Ratio}
Let $G_1 = \gauss{\mu_1}{\sigma_1^2}$ and $G_2 = \gauss{\mu_2}{\sigma_2^2}$ be two Gaussian distributions.  We compare $G_2$ to $G_1$ by evaluating the ratio of the likelihood of $G_2$ to $G_1$ given observations drawn from $G_1$.
We can do this empirically by drawing $N$ independent samples $y_1,\cdots,y_N$
from $G_1$, and evaluate the likelihood ratio $\LR{G_1}{G_2} :=
\prod_{i=1}^N \frac{p(y_i|G_2)}{p(y_i|G_1)} =\exp{\left\{-\sum_{i=1}^N
\log{\left(\frac{p(y_i|G_1)}{p(y_i|G_2)}\right)}\right\}}$.  Here $y_i$ are the
independent observations of $Y$ and $p(\cdot|G_2)$ is the likelihood function
of $G_2$.  We write the likelihood ratio as the exponential of
the negative sum of log-likelihood ratios to use the Kullback-Leibler (KL)
divergence to compute this in closed form, instead of drawing samples, i.e.,
\begin{align*}
\sum_{i=1}^N \log{p(y_i|G_j)} &\propto \frac{1}{N} \sum_{i=1}^N \log{p(y_i|G_j)}\\
&\stackrel{N\rightarrow\infty}{\approx} \E_{G_1}[\log[p(Y|G_j)]
\end{align*}
With this substitution,
\begin{align*}
\sum_{i=1}^{N} &\log{\frac{p(y_i|G_1)}{p(y_i|G_2)}} = 
\sum_{i=1}^{N} \log{p(y_i|G_1)} - \sum_{i=1}^{N} \log{p(y_i|G_2)}\\ &\propto \sum_{i=1}^{N} \frac{1}{N}\log{p(y_i|G_1)} - \frac{1}{N}\sum_{i=1}^{N} \log{p(y_i|G_2)}  \\
&\stackrel{N\rightarrow\infty}{\approx} \E_{G_1}[\log{p(Y|G_1)}] - \E_{G_1}[\log{p(Y|G_2)}]  \\
&= \KL{G_1}{G_2}
\end{align*}
the likelihood ratio becomes
$$
\LR{G_1}{G_2} = \exp{\big( - \KL{G_1}{G_2}  \big)},
$$
which we can evaluate in closed form for two Gaussian distributions $G_1$ and
$G_2$. Since $\KL{G_1}{G_2} \in [0,\infty)$ and is continuous
in all the parameters of $G_1$ and $G_2$, it follows that $\LR{G_1}{G_2} \in (0,1]$. Thus, we can interpret $LR{G_1}{G2}$ as the similarity
of $G_2$ and $G_1$. 
The likelihood ratio $\LR{G_1}{G_2}$ is a monotonic decreasing function of
$\KL{G_1}{G_2}$ and therefore not symmetric. In the comparisons we make, we set $G_1$ to be the predictive
distribution of the full GP, which we assume is ``correct'' and $G_2$ to be the
predictive distribution of the HGP. $\LR{G_1}{G_2}$ then tells us how well the
HGP models after the full GP.